\documentclass[smallcondensed]{svjour3}     
\smartqed  
\usepackage{graphicx}
\begin{document}

\title{Hippocampus Temporal Lobe Epilepsy Detection Using a Combination of Shape-based Features and Spherical Harmonics Representation
}


\author{Zohreh Kohan \and \\  
Hamidreza Farhidzadeh \and \\
        Reza Azmi\and \\
        Behrouz Gholizadeh 
}


\institute{Z.Kohan \and R.Azmi \and B.Gholizadeh \at
              Department of Computer Engineering, Alzahra University, Tehran, Iran \\
            \and
             *Corresponding Author :H.Farhidzadeh \at
              Department of Computer Science and Engineering, \\University of South Florida,\\ Tampa, FL, USA \\
             \email{hfarhidzadeh@mail.usf.edu}
}

\maketitle

\begin{abstract}
Most of the temporal lobe epilepsy detection
approaches are based on hippocampus deformation and use complicated features, resulting, detection is done with complicated features extraction and pre-processing task. In this paper, a new detection method based on shape-based features and spherical harmonics is proposed which can analysis the
hippocampus shape anomaly
and detection asymmetry. This
method consisted of two main parts; (1) shape feature extraction, and (2) image classification. For evaluation, HFH database is used which is publicly available in this field. Nine different geometry and 256 spherical harmonic features are introduced then selected Eighteen of them that detect the asymmetry in hippocampus significantly in a randomly selected subset of the dataset. Then a support vector machine (SVM) classifier was employed to classify the remaining images of the dataset to
normal and epileptic images using our selected
features. On a
dataset of 25 images, 12 images were used for feature extraction and the rest 13 for classification. The results show that the proposed method has accuracy, specificity and sensitivity of, respectively, 84\%, 100\%, and 80\%. Therefore, the proposed approach shows acceptable result and is straightforward also; complicated pre-processing steps were omitted compared to
other methods.
\keywords{temporal lobe epilepsy\and spherical harmonics\and hippocampus shape-based features\and classification\and SVM}
\end{abstract}

\section{Introduction}
\label{intro}
In some of the neurological diseases one or more brain components and substructures are affected. Some of these effects deform the shape of those components in one of the brain hemisphere or both. Therefore, unilateral and/or bilateral shape analysis of brain structures could lead to a better understanding of the abnormalities. There are evidences in the literature \cite{RefJ,Ref2,Ref3,Ref4} showing that the brain structures deform when some diseases such as schizophrenia, Alzheimer’s, Parkinson, and epilepsy occur. So using precise representation is helpful to parameterize the deformation that will useful to diagnosis of the mentioned illnesses automatically.
Hippocampus is a brain structure that belongs to the limbic system and is located in the medial temporal lobe. Hippocampus plays an important role in the formation of declarative, emotional, and long-term memories as well as language processing \cite{Ref5}. Hippocampus is one of the main targets of deformation in the brain in some disorders such as temporal lobe epilepsy. So, 3D representation and analysis of this structure could be useful in prognosis and diagnosis of those types of diseases.
To characterize the deformation of the brain structures volumetric \cite{Ref6} or shape analysis is used. Volumetric analysis is easy to interpret deformation but shape analysis is more common because of its ability in geometric and morphometric features description. There are two groups of approaches used for shape representation; explicit and implicit shape representation. In explicit representation, the shape is illustrated as a parametric form and in explicit representation it represents as the level set of a scalar function. There are some examples of shape analysis methods in the literature in which either implicit or explicit shape representation is used.
Statistical analysis of the anatomical shape deformations, which occur in epilepsy and other neurological disorders, require both global and local parameter based characterization of the anatomical shape that is deformed. Size- and volume- based analysis is the most popular method to achieve the parameterization of the shape deformations\cite{Ref8,Ref9}.
Many methods are proposed to analysis hippocampus are needed complicated pre-processing task. Most of them are reviewed in\cite{Ref8,Ref9,Ref10,Ref11}.
Since many of the shape-based features can illustrated hippocampus deformation and employ for detection. This features are straightforward and avoided complicated pre- processing steps.
This method utilized spherical harmonics (SPHARM) that is a powerful mathematically tool that used for representation and analysis of 3D closed surfaces \cite{Ref15}. SPHARM can be utilized in 3D representation and analysis of the brain structures. However, in contrast to previous studies on hippocampus shape analysis using SPHARM-Based shape descriptor, this work has been focused on the detection of asymmetry in hippocampus shape using SPHARM coefficients as the shape features. A sub set of the coefficients that were able to detect the asymmetry in left and right hippocampus was used to classify the images into either normal or epilepsy patient groups. Moreover, the complexity of this method is less compared to the other presented methods.
As a result, in this paper, Combination of shape-based features and spherical harmonics, this proposed method can analysis hippocampus shape and diagnosis temporal lobe epilepsy in MR images.
\section{PREVOIUS WORKS}
\label{sec:1}
Shen et al. \cite{Ref7} have reported about deformation of the hippocampus in epilepsy patients. The deformation of hippocampus has led the research groups to study the shape of the hippocampus for diagnosis purposes.
Gerig et al. \cite{Ref8} to analysis hippocampal shape use volume measurments and shape-based features. Shape-based feature that is used is based on Mean Square Distance (MSD) between left and right hippocampus surface shapes. Then Support Vector Machines (SVM) is employed to classify, also leave- one-out cross validation is applied to evaluation performance.
Csernansky et al. \cite{Ref9} used registration to compare the hippocampal volume and shape characteristics in patient and control subjects. Computing transformation vector from the points on the hippocampus surface represents shape.
Kodipaka et al. \cite{Ref10} used the Kernel Fisher Discriminant (KFD) algorithm for shape-based classification of hippocampal shapes. In this method, some landmarks manually placed on the hippocampus surface by an expert that indicated boundaries of the shape, then feature extraction is done by fitting a model to the landmarks using a deformable pedal surface, see detail about a deformable pedal surface in \cite{Ref13}. The result of this fitting process is a smooth surface that used in similarity alignment followed by a level-set non-rigid registration that describe in \cite{Ref14}. The output of this registration is local deformation between left and right hippocampus that used as input to the Kernel Fisher classifier.
Esmaeilzadeh et al. employed spherical harmonics (SPHARM) to analysis hippocampus shape \cite{Ref11}. This method alignment hippocampus left and right to each other the extract features based on spherical harmonic confidents, finally using SVM for calssification.
Shishegar et al. have used Laplace Beltrami operator for TLE diagnosis \cite{Ref12}. This method proposed a feature vector that described size and shape based on Laplace Beltrami eigenvalues and eigen-functions.Some of pervious methods need registration as pre- processing step \cite{Ref6,Ref7,Ref8}. Also in \cite{Ref10} using shape alone could not capture shape differences. Furthermore the most of reviewed methods is complicated and need pre-processing step.
In this paper, a new method is presented that using straightforward shape-based features for detection and omitted complicated pre-processing step such as registration.

\begin{figure}
  \includegraphics[scale=.7] {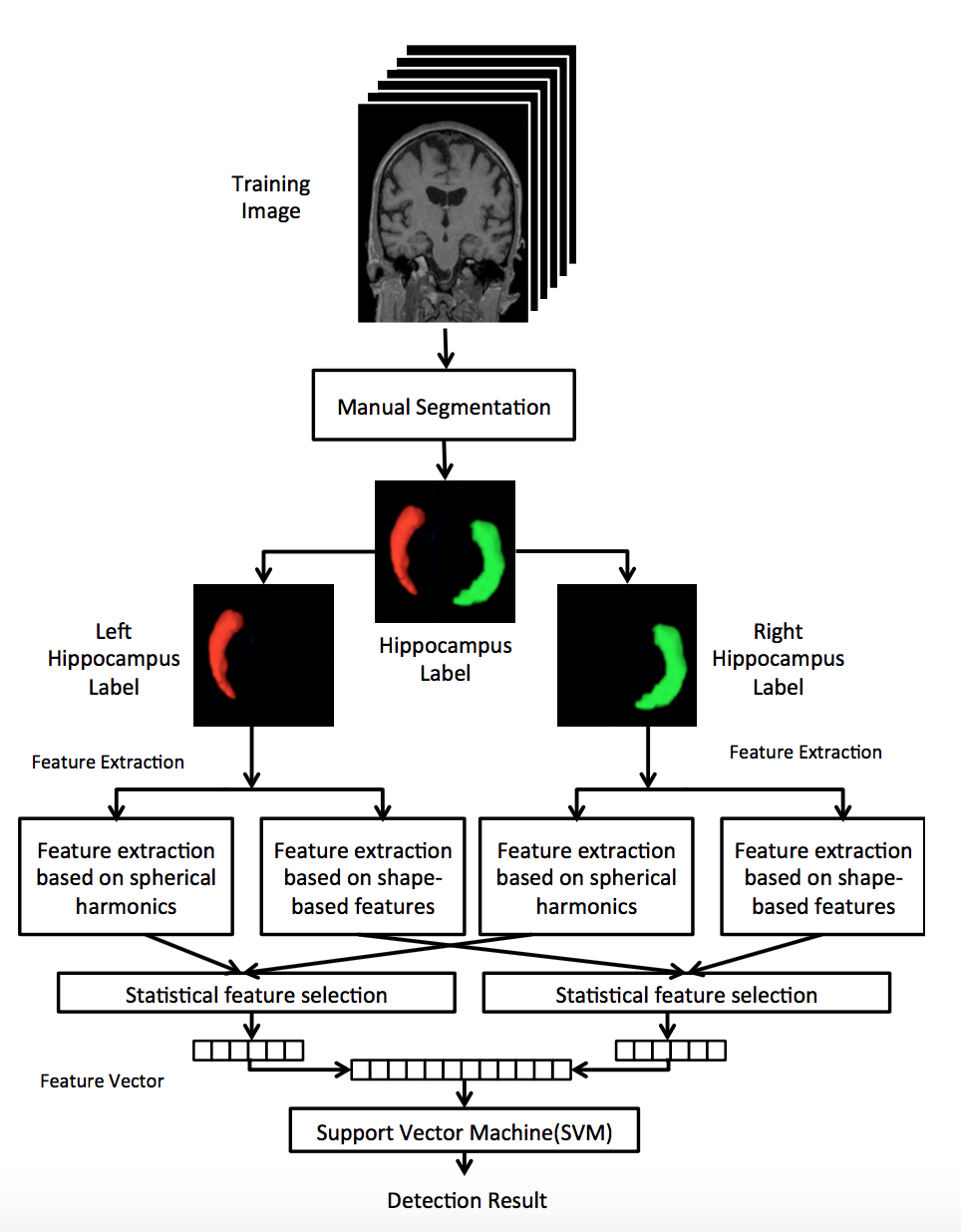}
\caption{Purposed method block diagram}
\label{fig:1}       
\end{figure}

\section{THE PROPOSED METHOD}
\label{sec:1}
The purposed method has two parts for feature extraction. In each part using statistical t-test most significant features is selected. Then a main feature vector is formed. Finally SVM is employed for classification hippocampus shapes. Fig.1 illustrate block diagram of the purposed method.

\subsection{Materials}
\label{sec:2}
Our dataset contained 25 T1-weighted coronal brain MR images (MRI) from Henry Ford hospital. MRI data is very useful for imaging in different applications such as breast cancer \cite{Ref28} and sarcomas \cite{Ref29}. 20 of the images were acquired from temporal lobe epilepsy subjects with medically intractable seizures and the remaining five were from normal subjects used as control dataset. Ten images out of 25 (all from epilepsy patients) were acquired with 1.5 Tesla MRI scanners and the remaining ones (15 images from epilepsy patients and all the normal cases) were from 3.0 Tesla MRI scanners. Slice thickness of all the images was 2.0 mm and the  {2015},in-plane resolution of the pixels varied from $0.39 \times 0.39 mm$ to $0.75\times 0.75mm$. For each image one expert provided a manual segmentation, then the segmentation labels were reviewed and confirmed by two more experts.
\subsection{Feature extraction based on spherical harmonics}
\label{sec:2}
This part consists of pre processing steps that explained below.
\subsection{Surface meshing}
\label{sec:2}
To generate a surface mesh for hippocampus shape, first, 3D manual segmentation was applied. For the further shape analysis and also for the classification of the hippocampus structure, it was necessary to have a mathematical representation of the 3D structure. One of the most common and simplest representations for explicit representation of the 3D surfaces is triangulation. In this method the shape is represented as a set of 3D discrete points that connected via triangular mesh. The marching cube algorithm \cite{Ref26} is a popular triangulation method applied to a 3D surface to obtain triangle surface mesh. After applying marching cube algorithm, the 3D shape is represented by coordinates of triangle mesh vertices as:
\begin{equation}
X=[x_1,y_1,z_1,...,x_n,y_n,z_n]^T
\end{equation}
Where n is the number of vertices.
\subsection{Spherical parametrization}
\label{sec:2}
First step in SPHARM representation is mapping the 3D surface to unit sphere under bijective mapping with lower distortion in area and topology called spherical parameterization \cite{Ref25}. This mapping is applied to a 3D surface represented as a triangular mesh. After mapping the surface mesh on the unit sphere each vertices can be represented in spherical polar coordinate with two parameters, the inclination angle $\theta$ and the azimuth angle $\varphi$ .

In this paper, for spherical parametrization we use, CALD algorithm that proposed by Shen and Makedon \cite{Ref26}. This algorithm consists of two steps: first, initial parametrization for triangular mesh \cite{Ref25} and second, local smoothing and global smoothing parametrization improve the quality of the parametrization. The goal of local smoothing step is minimization of the area distortion at a local sub mesh, this goal is achieved by solving a linear system and controlling its worst length distortion simultaneously. The global smoothing step as the second step of the algorithm compute the distribution of the surface distortions for all the mesh vertices, to equalize them over the complete sphere. In an overall view, the CALD algorithm merges the local and global methods together and executes each method alternately until a best parametrization is achieved. Fig.2 shows the left part of a hippocampus mapped to the unit sphere through this 3D mapping approach.
\subsection{SPHARM analysis}
\label{sec:2}
The SPHARM is a powerful mathematically tool that used for representation and analysis of 3D closed surface. The overall SPHARM analysis includes three main steps:
\begin{itemize}
\item {Surface meshing}
\item {Surface parametrization}
\item {Spherical expansion}
\end{itemize}
\begin{figure}
  \includegraphics[scale=0.75]{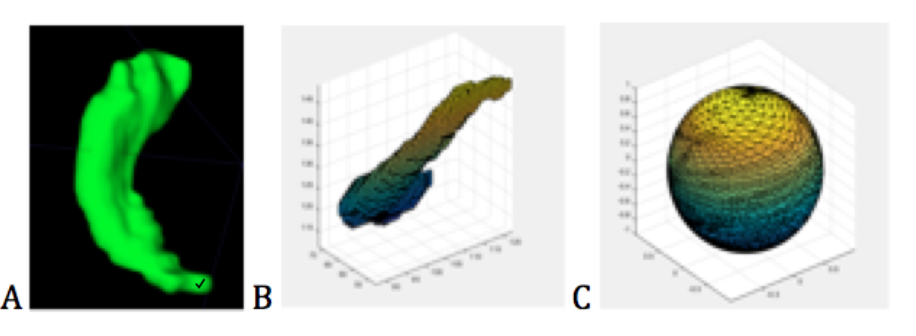}
\caption{A) Left hippocampus. B) The result of marching cube meshing algorithm. C) The result of spherical parameterization}
\label{fig:1}       
\end{figure}

The first two steps have been explained in section II.C. When the 3D surface is represented via two parameters (the inclination angle $\theta$ and the azimuth angle $\varphi$) the surface
can be expanded into a complete set of SPHARM basis functions $Y^m_l$ .

\begin{equation}
Y_l^m(\theta,\varphi)=\sqrt{\frac{2l+l(1-m)!}{4\Pi(1+m)!}}p_l^mcos(\theta)e^{im\varphi}
\end{equation}

where
\\
$$0<l<L_{max}$$
\\
and
\\
$$-l<m<l$$
\\

and $p_l^mcos(\theta)$  is the associated Legendre polynomials defined
by the differential equation as follows:

\begin{equation}
 P_l^m(x)=\frac{(-1)^m}{(2^il!)}(1+x^2)^{m/2}\frac{(d^{l+m})}{(dx^{l+m})}(x^2-1)^l
\end{equation}
 The surface is independently decomposed through SPHARM as:
 
 \begin{equation}
x(\theta,\varphi)=\sum_{l=0}^{L_max}{\sum_{m=-l}^{l}{C_{lx}^mY_l^m(\theta,\varphi)}}
\end{equation}

\begin{equation}
y(\theta,\varphi)=\sum_{l=0}^{L_max}{\sum_{m=-l}^{l}{C_{ly}^mY_l^m(\theta,\varphi)}}
\end{equation}

\begin{equation}
 z(\theta,\varphi)=\sum_{l=0}^{L_max}{\sum_{m=-l}^{l}{C_{lz}^mY_l^m(\theta,\varphi)}}
\end{equation}

These terms can be combined into a single function:
\begin{equation}
v(\theta,\varphi)=\sum_{l=0}^{L_max}{\sum_{m=-l}^{l}{C_l^mY_l^m(\theta,\varphi)}}
\end{equation}

Where
\begin{equation}
v(\theta,\varphi)=(x(\theta,\varphi),y(\theta,\varphi),z(\theta,\varphi))^T
\end{equation}

And

\begin{equation}
C_l^m=(C_{lx}^m,C_{ly}^m,C_{lz}^m)^T
\end{equation}
The coefficients $C_l^m$ are computed using least-squares estimations. According to equation (7) these coefficients are determined up to user-desired maximum degree $L_max$. 

The original surface is indicated as $X=(x_1,x_2,x_3,...,x_n)^T$  and  $(a_1,a_2,a_3,...,a_k)$ is an estimate for the coefficients which are
Featureselection
obtained by solving previous equation follow manner:
\begin{equation}
C_l^m\cong{{Y_l^m}^TY_l^m}^{-1}X
\end{equation}
As claimed by equation (7), after coefficients are estimated, the 3D surface can be reconstructed. Fig.3 illustrates reconstruct shape in four different degrees using higher degrees yield to more details in the reconstructed shape.

\begin{figure}
  \includegraphics[scale=0.8]{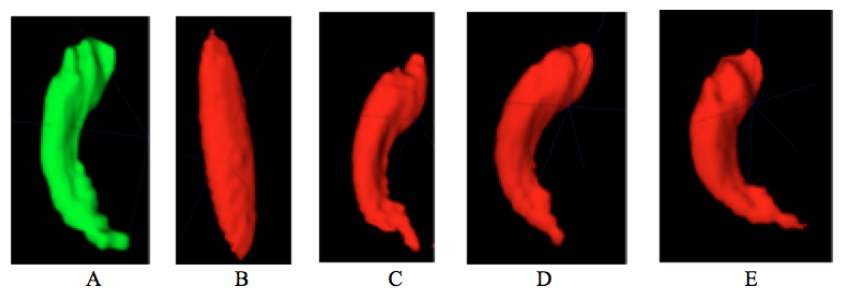}
\caption{ A) 3D shape of left hippocampus and (B-E) its reconstruction through SPHARM based shape reconstruction using (B) $L-max$=1 (C) $L-max$ =8 (D) $L-max$ =16 and (E) $L-max$ =24}
\label{fig:1}       
\end{figure} 

\begin{figure}
  \includegraphics[scale=0.7]{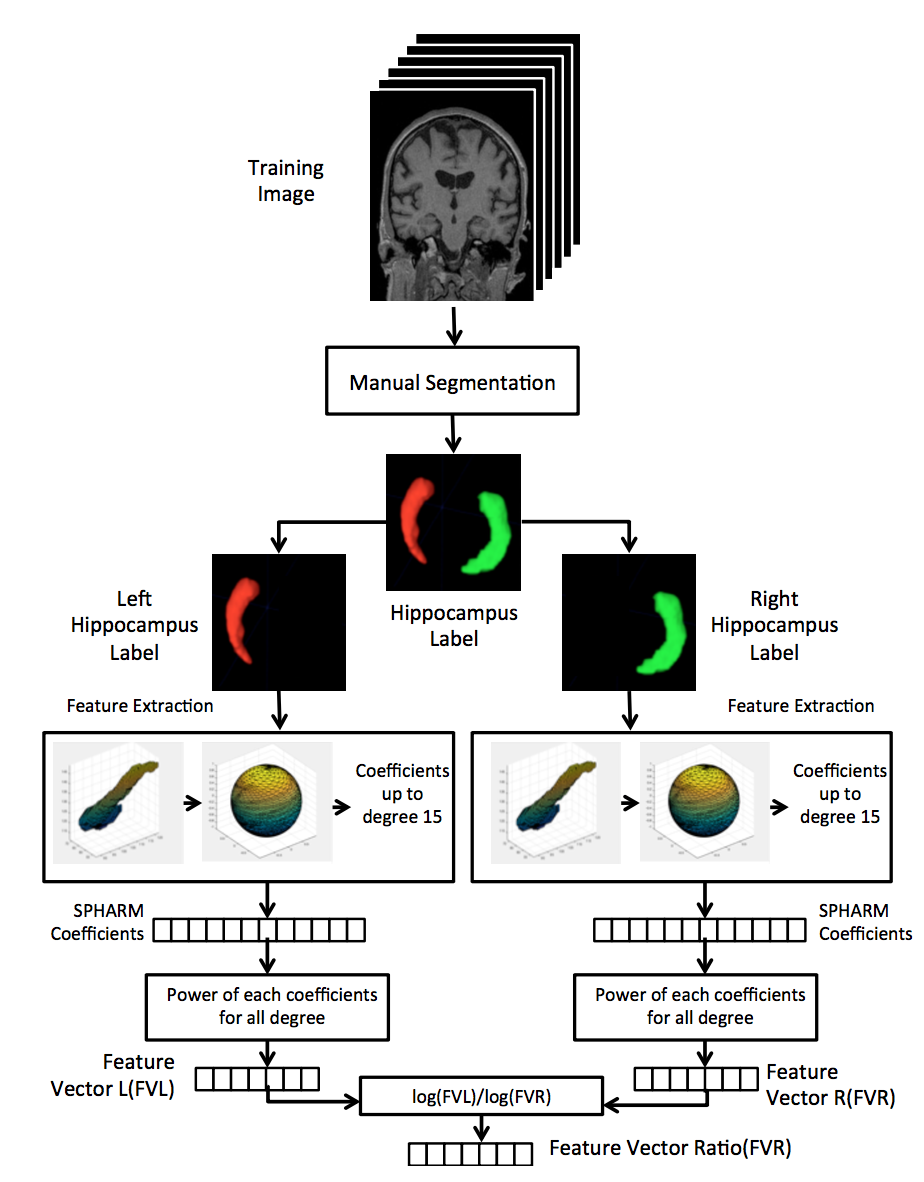}
\caption{ Feature extraction}
\label{fig:1}       
\end{figure} 

\subsection{Featureselection}
\label{sec:2}
The real parts of SPHARM coefficients were used as features in this work. For each subject hippocampus has two parts; left and right. For each part the coefficients were computed up to degree 15. So the length of feature vector is
$3\times(15+1)^2 =768$ for each side. To make the coefficients comparable across left and right hippocampus, we needed to
normalize them by cancelling out the translational and rotational misalignments. Rotation was achieved by alignment of reconstructed shape based on first degree coefficients only (i.e. an ellipse). We used the alignment algorithm that used in \cite{Ref26}. For that purpose we aligned the shortest and the longest axis of the ellipse along x-axis and z-axis, respectively. We also translated the shapes to the origin of the coordinate system by ignoring the coefficients of degree 0. Since we used the ratio of left and right coefficients for each hippocampus, no scaling adjustment was required.
After normalizing the coefficients, we calculated sum of the power of each coefficients for all the degrees as shown in equation (11).

\begin{equation}
(\sqrt{\sum_{m<|l|}{||C_0^m||^2}},\sqrt{\sum_{m<|l|}{||C_1^m||^2}},\sqrt{\sum_{m<|l|}{||C_2^m||^2}},...,\sqrt{\sum_{m<|l|}{||C_l^m||^2}})
\end{equation}

This step yielded to a feature vector of $(15+1)^2 =256$ elements. Since our aim was to detect hippocampal asymmetry
we calculated the ratio of logarithm of left coefficients to logarithm of right coefficients. We used logarithm to makes the features of different orders more comparable.(Fig.4)

\begin{figure}
  \includegraphics[scale=0.7]{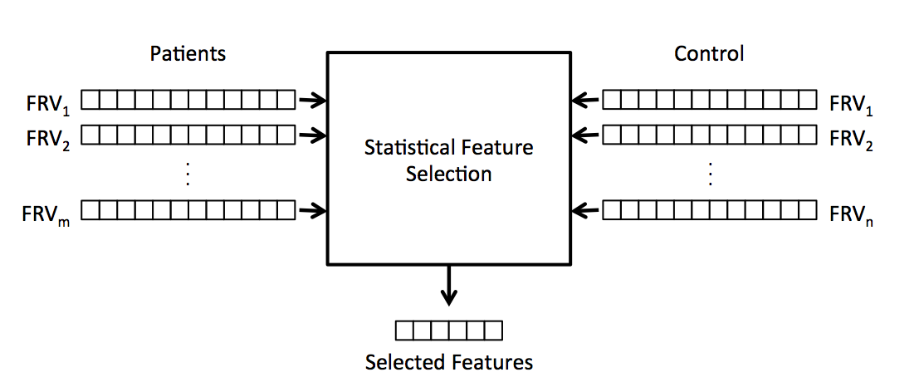}
\caption{ Feature extraction}
\label{fig:1}       
\end{figure} 

Since only some of the listed features carried applicable information for the classification of shape (significantly differentiate between normal and abnormal shapes), we employed a statistical t-test to identify and select the most discriminative features. Fig.5 shows a schematic diagram of the feature selection algorithm. We run t-test on the each element (feature) of features vector to compare the element values of patients and normal subjects in a training set. For two sets of samples (in this paper one set is the feature values for normal cases and the second set is the feature values for the epilepsy cases) with averages of M1 and M2 the p-value indicates the probability that the observed difference between M1 and M2 is due to chance under the null hypothesis that M1 and M2 are the same. Therefore, a lower p-value shows statistically stronger difference that corresponds to a more significant feature. We obtained the associated p-value for each feature and selected those features with p-values of less than 0.05 as significantly discriminative features.
Our hypothesis was that the significant features could help better in classification of hippocampus into healthy and epileptic classes. To select the best features that are able to differentiate between normal and epileptic subjects, first we randomly selected a subset of 12 images from our data set to form our training set. The training set contained both normal (two images) and epileptic (ten images) subjects. The remaining images were used as the test dataset in the classification explained in the next subsection. We extracted all SPHARM coefficients then created the feature vectors as described; we used t-test to select those features in which the value of the feature between normal and patient cases was detected significantly different ($p-value < 0.05$).
\begin{table}
\caption{Shape-based features and corresponding p-values}
\label{tab:1}       
\begin{tabular}{lll}
\hline\noalign{\smallskip}
Feature & p-value  \\
\noalign{\smallskip}\hline\noalign{\smallskip}
Maximum shape diameter &	0.7187& \\
\hline
Shape volume	 & $2.54 \times 10^{-4} $** &\\
Surface area & 0.7690 &\\
Compactness\cite{Ref26} &0.7054 &\\
Mesh size & 0.0015** &\\
First-order border moments\cite{Ref25} & $2.38 \times 10^{-4} $ **&\\
Third-order border moments\cite{Ref25} & 0.5088 &\\
Circumscribed sphere volume to &  0.6028 &\\
shape volume ratio & &\\
Curvature & 0.075* &\\ [1ex] 
\noalign{\smallskip}\hline
\end{tabular}
\end{table}

\subsection{Feature extarction based on shape-based features}
\label{sec:2}
In this section we first introduced all the features we used to study shape characteristics.
Table 1 shows a list of shape and size features we have used in this project to detect asymmetry between right and left hippocampus. Fig.6 shows a general block diagram of our feature extraction method. Following we described these features.

\begin{figure}
  \includegraphics{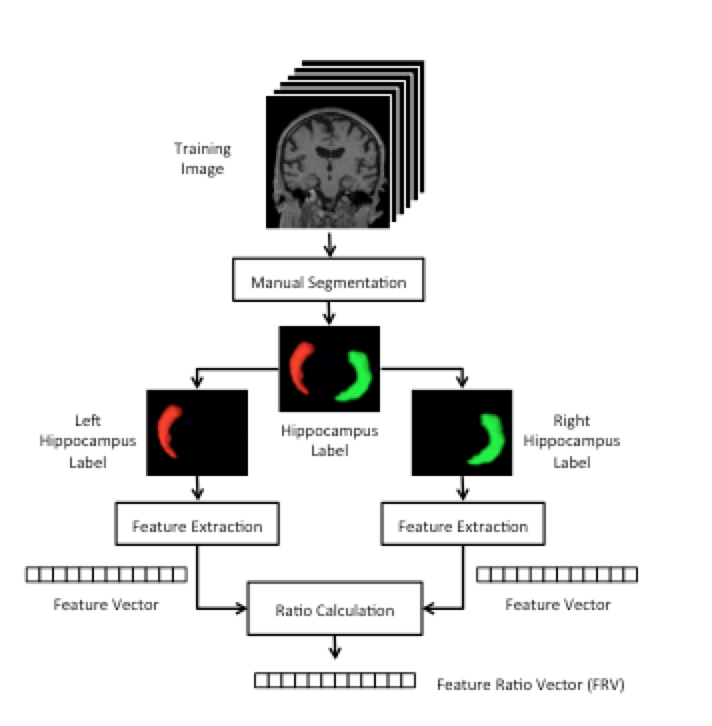}
\caption{ Feature extraction}
\label{fig:1}       
\end{figure} 

\subsubsection{Maximum shape diameter}
\label{sec:2}
The maximum possible Euclidian distance between two points of the surface of the shape. Since epilepsy disease could deform the surface of the hippocampus \cite{Ref25} it might changes maximum diameter of the shape.
\subsubsection{Shape volume}
\label{sec:2} To measure the shape volume of right and left hippocampus we multiply number of voxels by the volume of a single voxel.
Surface area: To measure the surface area we calculated the summation of the outer surfaces of all the voxels that formed the hippocampus surface.
\subsubsection{3D Compactness}
\label{sec:2} Compactness is a unit less shape feature that describes how closely packed the shape is. In the other hand, it shows the roughness of a shape surface to its volume \cite{Ref26}. Sphere has the smallest value of 3D compactness i.e. about 113. As the sphere deforms towards a more complicated shape, compactness becomes larger and larger. 3D compactness (C) calculated as ratio of the cubed surface area (A) to the squared surface volume (V):

\begin{equation}
C=\frac{A^3}{V^2}
\end{equation}

\subsubsection{Mesh Size}
\label{sec:2}In this paper we measured the number of vertices in the shape surface mesh using marching cube meshing algorithm \cite{Ref26}. The marching cubes algorithm is a 3D iso-surface representation technique for generating mesh for a 3D surface.

\subsubsection{Moments}
\label{sec:2} Moments are the statistical property of the shape that are evolved from the moment concept in physics. There are two types of moments, region moments and boundary moments. In this work, we use boundary moments \cite{Ref25}. The pth moment is defined as:

\begin{equation}
m_p=\frac{1}{N}{\sum_{i=1}^{N}{[z(i)]^p}}
\end{equation}

where d(i) is Euclidian distance between ith voxels and the centroid, N is number of voxels.
The pth central moment defined as:

\begin{equation}
M_p=\frac{1}{N}{\sum_{i=1}^{N}{[z(i)-m_1]^p}}
\end{equation}

In this paper we use low-order moments such as M1
and M3 that are defined as below:
\begin{equation}
F_1=\frac{[\frac{1}{N}{\sum_{i=1}^{N}{[z(i)-m_1]^2}}]^{\frac{1}{2}}}{\frac{1}{N}{\sum_{i=1}^{N}{z(i)}}}
\end{equation}
and

\begin{equation}
F_3=\frac{[\frac{1}{N}{\sum_{i=1}^{N}{[z(i)-m_1]^4}}]^{\frac{1}{4}}}{\frac{1}{N}{\sum_{i=1}^{N}{z(i)}}}
\end{equation}
where

\begin{equation}
m_1=\frac{1}{N}{\sum_{i=1}^{N}{z(i)}}
\end{equation}

\subsubsection{Circumscribed sphere volume to shape volume ratio}
\label{sec:2} This is the ratio of the volume of the circumscribed sphere centred at the shape centroid to the volume of the shape.
\subsubsection{Curvature}
\label{sec:2} We defined the curvature of the shape as follows:

\begin{equation}
C=\frac{d_{C1}+d_{C2}}{d_{12}}
\end{equation}

Where $d_C1$ and $d_C2$ are the Euclidian distances between centroid of the shape and two opposite tips of the shape that are located in the maximum distances to the centroid and $d_12$: is the Euclidian distance between the two tips of the shape.
To analysis the shape and size of the left and right hippocampus, we extracted all the features for left and right hippocampus, separately. Then we measured the ratio of right feature values to left feature values as well as left feature values to right. We chose the ratio that is greater or equal to 1 as an indicator for hippocampal symmetry/asymmetry.
In this part feature selection is done according to statistical feature selection test that describe in Feature Selection part.
\subsection{Classification}
\label{sec:2} 
Support vector machine (SVM) is a supervised learning models used for classification and regression analysis with works very well on similar studies \cite{Ref27,Ref28,Ref29}. Linear SVMs search for the optimal hyperplane that separate the dataset to groups, i.e. the hyperplane that makes the maximal margin between groups. For more information about SVM classifiers you can see \cite{Ref26}. We used SVM classifier to classify each test image to normal or patient categories using the selected features (Fig.7). We used our test dataset to evaluate the performance of our classification algorithm.
\subsection{Evaluation metrics}
\label{sec:2} 
To evaluate our detection algorithm performance we calculated specificity, sensitivity and accuracy metrics (equations 19 to 21) for the classifier results. These metrics have been largely used in medical image analysis \cite{Ref25,Ref26}.

\begin{equation}
Accuracy=\frac{TP+TN}{TP+FN+TN+FP}\times100
\end{equation}

\begin{equation}
Sensitivity=\frac{TP}{TP+FN}\times100
\end{equation}

\begin{equation}
Specificity=\frac{TN}{FP+TN}\times100
\end{equation}

TP , TN , FP and FN are, respectively, true positive, true negative, false positive and false negative.
\begin{figure}
  \includegraphics[scale=0.9]{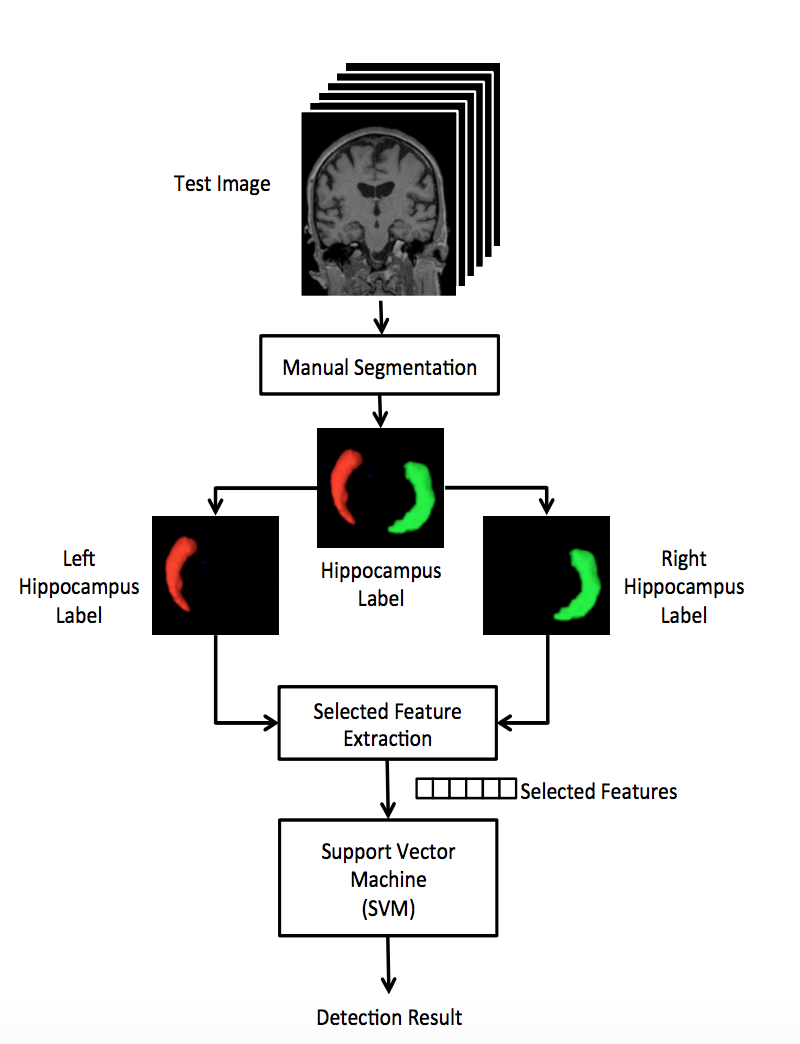}
\caption{ Classification block diagram}
\label{fig:1}       
\end{figure} 

\subsection{Experimental results}
\label{sec:2}
We run our algorithm on 25 segmentation labels of hippocampus on T1-weighted MRI. We divided our image dataset to two subsets; training and test. Training images (12 images, 2 normal and 10 epileptic) were used for feature selection. Test image set consisted of 13 images; 3 normal and 10 epileptic images.
All the implementations of this work were implemented in MATLAB platform and run on a Mac operating system. 
\subsubsection{Features selection}
\label{sec:2}
Feature selection consists of two parts. First, feature extraction based on spherical harmonics coefficients. Second, feature extraction based on shape-based features. In each part statistical t-test is used for feature selection. In both parts, the most significant features (associated with p-value $<$ 0.05) is selected. Then combination two selected features create feature vector.
\subsubsection{Classification}
\label{sec:2}
Leave–one–out cross validation methodology has been applied to divide the test image set to one test image and 12 training for SVM classifier. Having the 18 selected features used in our SVM classifier, the specificity, sensitivity, and accuracy of our detection algorithm were 84\%, 100\% and 80\%, respectively.
\subsubsection{Conclusion and future work}
\label{sec:2}
Our preparatory results show that combination SPHARM coefficients and shape-based features could be helpful to describe the hippocampus shape deformation and could be used in diagnosis of the temporal lobe epilepsy disease in MRI. Since the dataset we used in our study is formed based on some part of the dataset used in \cite{Ref10,Ref11}, not the whole, it is challenging to reach a fair comparison between presented results and ours. As an advantage, the complexity of our algorithm is less than the other presented algorithms. In contrast to other work, our method detects the asymmetry in hippocampus structure shape for diagnosis of epilepsy and the preliminary results showed that hippocampal asymmetry detection could be useful for diagnosis of the brain abnormalities such as temporal lobe epilepsy. However, small size of the dataset is a limitation in our work.
As future work we are going to increase the number of images in the training and test dataset. We will also compare the results to other reported results in the literature using appropriate statistical tests to examine the significance of the results. We would like to improve the results of our work by adding more shape-based features that are able to significantly separate patients from control groups. It could be also considered to use and evaluate this algorithm for detecting other brain abnormalities.




\end{document}